\documentclass[letterpaper, 10 pt, conference]{ieeeconf}  % Comment this line out if you need a4paper

\IEEEoverridecommandlockouts                              % This command is only needed if 
\usepackage{epsfig} % for postscript graphics files
\usepackage{amsmath} % assumes amsmath package installed
\usepackage{amssymb}  % assumes amsmath package installed
\usepackage[linesnumbered,ruled]{algorithm2e}
\usepackage{booktabs}
\usepackage{graphicx}
\usepackage{array}
\usepackage{soul}
\usepackage{color, xcolor}
\soulregister{\cite}7 % 注册\cite命令
\usepackage{multirow}
\usepackage{makecell}
\usepackage{balance}

\title{\LARGE \bf
DB3D-L: Depth-aware BEV Feature Transformation for Accurate 3D Lane Detection
}

\author{Yehao Liu, Xiaosu Xu, Zijian Wang, Yiqing Yao, ~\IEEEmembership{Member,~IEEE} % <-this % stops a space
\thanks{This work is supported in part by the Funds for Creative Research Groups of China under Grant 61921004, in part by 2022 Ministry of Education's “Chunhui Jihua” Cooperative Research Project under Grant HZKY20220128, in part by Jiangsu Provincial Department of Science and Technology (No. BM2023013). \emph{(Corresponding author: Xiaosu Xu)}}% <-this % stops a space 
\thanks{The authors are with the School of Instrument Science and Engineering, the Key Laboratory of Micro-Inertial Instrument and Advanced Navigation Technology, Ministry of Education, Southeast University, Nanjing 210096, China (e-mail: liuyehao@seu.edu.cn; xxs@seu.edu.cn; 230218957@seu.edu.cn; yucia@sina.com).
}}

\UseRawInputEncoding
\begin{document}

\maketitle
\thispagestyle{empty}
\pagestyle{empty}

%%%%%%%%%%%%%%%%%%%%%%%%%%%%%%%%%%%%%%%%%%%%%%%%%%%%%%%%%%%%%%%%%%%%%%%%%%%%%%%%
\begin{abstract}

3D Lane detection plays an important role in autonomous driving. Recent advances primarily build Bird’s-Eye-View (BEV) feature from front-view (FV) images to perceive 3D information of Lane more effectively. However, constructing accurate BEV information from FV image is limited due to the lacking of depth information, causing previous works often rely heavily on the assumption of a flat ground plane. Leveraging monocular depth estimation to assist in constructing BEV features is less constrained, but existing methods struggle to effectively integrate the two tasks. To address the above issue, in this paper, an accurate 3D lane detection method based on depth-aware BEV feature transtormation is proposed. In detail, an effective feature extraction module is designed, in which a Depth Net is integrated to obtain the vital depth information for 3D perception, thereby simplifying the complexity of view transformation. Subquently a feature reduce module is proposed to reduce height dimension of FV features and depth features, thereby enables effective fusion of crucial FV features and depth features. Then a fusion module is designed to build BEV feature from prime FV feature and depth information. The proposed method performs comparably with state-of-the-art methods on both synthetic Apollo, realistic OpenLane datasets.

\end{abstract}

\vspace*{\baselineskip}  %~\\


%%%%%%%%%%%%%%%%%%%%%%%%%%%%%%%%%%%%%%%%%%%%%%%%%%%%%%%%%%%%%%%%%%%%%%%%%%%%%%%%
\section{INTRODUCTION}

Lane detection plays an important role in autonomous driving, such as trajectory planning and lane keeping \cite{yan2022once,guo2020gen,abualsaud2021laneaf,he2022lanematch,sun2022adaptive}. Lane detection comprises image space lane detection and three-dimensional (3D) space lane detection. Detecting lane lines (or points) in image space is straightforward, with some studies achieving high performance \cite{yan2022once,tabelini2021keep,ran2023flamnet,tang2021review,qin2020ultra}, while detecting the 3D spatial position information of lane lines (or points) is challenging. In the application of autonomous driving, 3D lane detection is  more needed to realize some downstream planning and control tasks.

With the widespread adoption of deep learning in computer vision, it has become feasible to detect 3D lane lines using a straightforward monocular camera. The objective of 3D lane detection is to generate the 3D spatial representation of lanes from input image(s). Existing methods can be classified into two-stage and end-to-end approaches based on whether they rely on the detection of 2D lane lines.
The two-stage methods firstly detect the 2D pixel coordinates in the image space. Then the 2D pixel coordinates is transferred to BEV space by inverse perspective mapping (IPM) \cite{jeong2016adaptive}, then 3D coordinates are recovered based on the ground plane assumption. Although these methods has a high precision in detecting 2D Lane lines (or points), it is restrained to recover 3D coordinates from the pixel coordinates of lane points by ground plane assumption in special sence, especially when the road is undulating \cite{guo2020gen, garnett20193d, wang2023bev}. In addition, the precision of 3D coordinates is limitated by the lack of camera's extrinsics and intrinsics.

End-to-end approaches leverage annotated 3D lane information for supervised training, aiming to detect 3D lane coordinates directly from input images. This approach mitigates the challenges associated with transforming information from 2D to 3D space. Additionally, to enhance the efficiency of representing 3D lane lines and establish a more effective link with downstream planning and control tasks, some methods often predict the 3D structure of lane lines directly from BEV space. Due to construct BEV maps from monocular images is typically a complex process, the recent methods consider incorporating camera ensemble information into the model design to enhance the performance of lane detection. However, the lacking of the depth information leads to a precision loss between the transformation from FV space to BEV space. \cite{philion2020lift,li2022bevformer,yang2023bevheight,yang2023bevheight++}. 

\begin{figure}[t]
	\centering
	\includegraphics[width=\columnwidth]{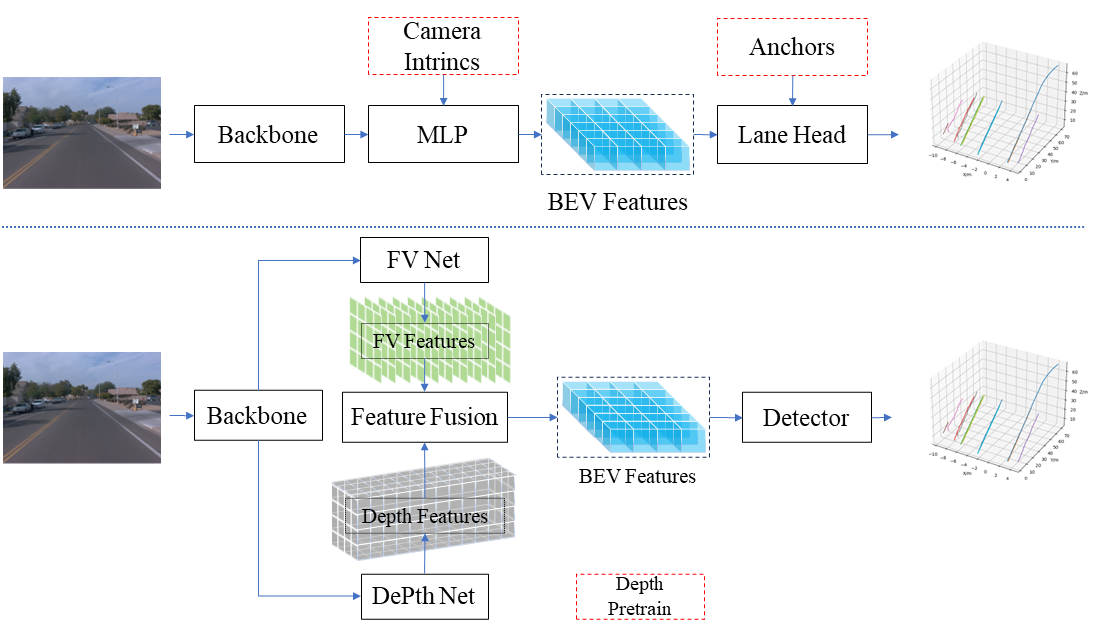}
	\caption{The pipeline of previous BEV lane detection (upper) and our proposed DB3D-L (lower). Previous methods mainly utilize camera parameters and inverse perspective mapping (IPM) to transform the features into BEV space. Our methods build BEV by depth feature and FV features fusion with attention mechanism. Note that only standard operators are adopted in DB3D-L.}
	\label{fig:intro}
\end{figure}

In addition, the process of transformation from FV to BEV has some issues. Transformer based methods use Vision Transformer (ViT) capture the BEV information from 2D images \cite{li2022bevformer,chen2022persformer,luo2023latr}. Although the use of ViT can achieve relatively high accuracy, its model is complex and difficult to deploy. LSS paradigm \cite{philion2020lift,li2023bevdepth,zhou2023matrixvt} creates BEV feature in a bottom-up approach. These methods generate a large and sparse index matrix to map the transformation from FV to BEV, incurring expensive computational costs.

Firstly, to overcome these limitations of current methods, we propose an depth-aware 3D lane detection method based on prime FV feature and depth fusion. To capture essential depth information in monocular images, a refined Depth Net is integrated into the proposed model. Secondly, an feature reduce method is introduced. In whichthe height dimension into single channel, as in our method, the key information we focus on does not overlap in the height dimension, there is only one instance at a given BEV grid cell (see Sec. 3.B). Finally, to effectively fuse the FV feature and depth feature, a fusion module with spatial attention mechanism is proposed, through which the BEV features are generated.
 
To summarize, the contributions of this study are fourfold:
\begin{itemize}
\item We propose DB3D-L, an end-to-end 3D Lane detection framework with an effective space transformation from FV space to BEV space. By integrated a Depth Net, DB3D-L creats BEV feature more in line with space geomtry.
% Different from exsiting methods, DB3D-L performs more robust and less rely on ground-plane assumption or camera intrincs while remains the advantage of disployment-friendly.

\item Feature reduce module PFE and DAT are proposed to reduce height dimension of FV features and depth features in DB3D-L, avoiding computations on ineffective information from elevated areas, thereby enables effective fusion of crucial FV features and depth features.
%We introduce an refined backbone that can both extract FV features and depth information, a SE layer is combined to learn spatial cross attention. 
\item A FusionNet is proposed, in which depth is allocated to related FV feature using attention mechanism in which BEV feature is fused effectively.

\item Extensive experiments demonstrate that DB3D-L yields comparable performance to the state-of-the-art method on the both synthetic Apollo, realistic OpenLane datasets while being more efficient and generally applicable.
\end{itemize}

\section{RELATED WORKS}

%In this section, previous work will be presented in three categories, namely correspondence matching based methods, keypoint detection methods, and partial-to-partial registration methods.

\subsection{Monocular depth estimation}

Monocular vision is wildly used in robotics systems \cite{bhoi2019monocular}, which plays a crucial role in various robotics and autonomous system applications, including ego-motion estimation, scene understanding, and obstacle avoidance. However, 3D reconstruction from monocular image is an ill-posed issue because the scale information is missing. Depth estimation refers to the process of estimating a dense depth map from the corresponding input image(s), which can recover the scale of the image, and the recovery depth information can be utilized to infer the 3D structure.

Learning based methods considered monocular depth estimation as a regression of a dense depth map from a single RGB image. DAV \cite{huynh2020guiding} uses a standard encoder-decoder scheme and proposes to exploit co-planarity of objects in the scene via attention at the bottleneck. Pseudo-LiDAR \cite{wang2019pseudo} reconstructs the point cloud from a single RGB image with off-the-shelf depth prediction networks, led to the recent advances in 3D detection. Adabins \cite{bhat2021adabins} contains a transformerbased architecture block that divides the depth range into bins whose center value is estimated adaptively per image. The final depth values are estimated as linear combinations of the bin centers.

\subsection{BEV Feature Transformation}

The spatial transformation (ST) is used to create BEV feature (see Fig. 2 (b) ) from FV features (see Fig. 2 (a) ). Which is the key of BEV-based spatial perception tasks.

LSS paradigm \cite{philion2020lift,li2023bevdepth,zhou2023matrixvt}  perform the ST in two stage, firstly a frustum of the image pixel is created by a depth estimation net, from which the 2D features is transferred to the 3D space, then it finds the mapping between the elements in frustum and the elements in BEV grids, using a mapping matrix is built in advance. All features in the frustum are ranked according to their corresponding BEV grid ID, and a cumulative summation is performed over all features to get the BEV featurures.

BEVFormer \cite{li2022bevformer} creates BEV feature by Vision Transformer (ViT). The ViT is typically suitable to conduct ST by its global attention machanism. In these method, an explicit learnable BEV Queries is Pre-generated and is used to capture the BEV information from muti-view 2D images. An explicit BEV feature is well suited to incorporate temporal information or features from other modalities and to support more sensing tasks simultaneously.

HeightFormer \cite{wu2023heightformer} creats BEV representation by modeling height in BEV space in an explicit way, rather than modeling depth in image views implicitly. In this network, heights and uncertainties are predicted in a selfrecursive way without any extra data aidded, reaching more rebust results.

Translateing images into maps \cite{saha2022translating} finds that regardless of depth, pixels in the same column on the image are along the same ray under BEV. Therefore, each column can be transformed to BEV to construct BEV feature maps. The authors encode each column in the image as memory, and query memory using the radius of the ray under BEV to obtain BEV features. Finally, the model has a better view conversion ability through data supervision.

\subsection{3D Lane Detection in Bird’s-Eye-View}

As dissussion in Section I, State-of-the-art methods tend to predict the 3D structureof lane lines directly from BEV space. These methods first transform an image view feature map into BEV, and then detect lane lines based on the BEV feature map. 3DlaneNet \cite{garnett20193d} utilizes IPM to transform features from FV into BEV and then regresses the anchor offsets of lanes in the BEV space. Gen-LaneNet \cite{guo2020gen} contains an extensible two-stage framework that separates the image segmentation subnetwork and the geometry encoding subnetwork. PersFormer \cite{chen2022persformer} contains a unified 2D and 3D lane detection framework and introduced transformers into the spatial transformation module to obtain more robust features. BEV-Lanedet \cite{wang2023bev} creats muti-layer features, multiple View Relation Module (VRM) is used for each layer of different sizes to realize dynamic mapping from FV to BEV, but the counts of  FPN layer and size is manual preset, which is limited for different senerios. These methods deploy intra-network feature maps IPM with camera intrinsic/extrinsic parameters, implicitly or explicitly.

\begin{figure}[t]
	\centering
	%	 	\framebox{\parbox{3in}{
	\includegraphics[width=\columnwidth]{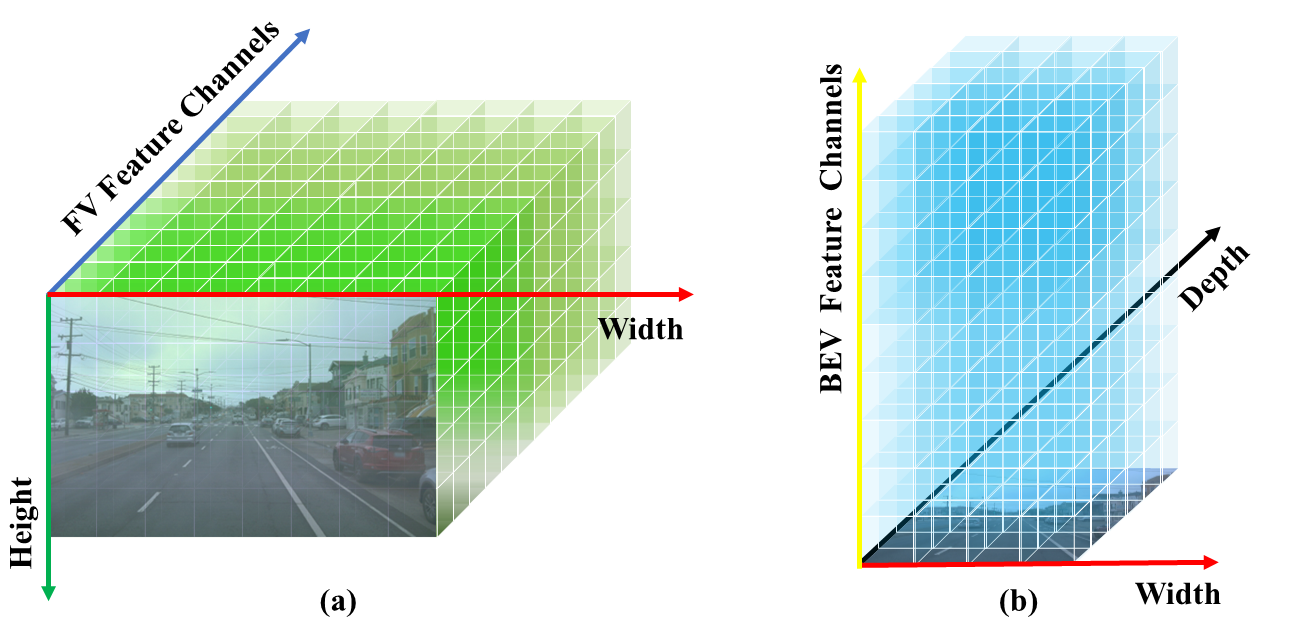}
	%			 	}}
	\caption{FV features (a) vs BEV features (b). We find that BEV features and FV features have little difference essentially, they are both N channels   2-dimension value map. So we build BEV features just as a features map of CNN outputs. The FV features and BEV features has the common dimension in width, which is critical for the transformation from FV to BEV.}
	\label{fig:features}
\end{figure}

\section{PROPOSED METHOD}

 Monocular 3D lane detection aims to predict the 3D position of lane points in camera cordinates for each lane line from an input image $\boldsymbol{I} \in  \mathbb{R}^{ H_I \times W_I \times 3}$ , The lane line is represented as $\boldsymbol{L}_k = \{pt _i\in \mathbb{R}^3 \mid i=1,2 , \ldots , N_k\}, k=1,2,...,K $, where $ K $ is the counts of total lane lines in the image, and $N_k$ is number of points in lane line $ L_k $. A comprehensive schematic of the entire workflow is shown in Fig. \ref{fig:pipeline}. The proposed DB3D-L network mainly consits of three modules: a refined backbone (D\&F Net) that  extracts depth information and FV features simultaneously, a feature fusion module (FusionNet) that builds BEV features from depth and  FV features, a 3D lane detection head (D-L Net) that outputs the 3D lane detection results from the BEV features.

\subsection{D\&F Net}

In order to be depth-aware, we estimate the depth of the input image and extracts FV features simultaneously at the beginning of the framework. Different from LSS paradigm that creats a frustum for the image, the depth is used for recovering a pseudo point clouds in 3D space of the sence in size of  $H_I \times W_I \times D$, where $H_I$, $W_I$ represent the height, width in image view respectively, $ D $ represent the channel of depth, in which considering the distortion caused by pespective mapping implicitly. As shown in Fig. \ref{fig:reduce}, the point clouds contains the correlation between FV and BEV. Both are projections of pseudo point clouds onto a plane. The difference lies in BEV being a direct projection of the 3D point clouds onto the ground plane, while the FV is obtained through perspective imaging. Therefore, besides obtaining BEV through an IPM from the frontal view, it is also possible to recover 3D point clouds from the frontal view and project them onto a top-down plane to achieve the same BEV.

\begin{figure*}[t]
	\centering
	% 	\framebox{\parbox{3in}{
	\includegraphics[scale=0.45]{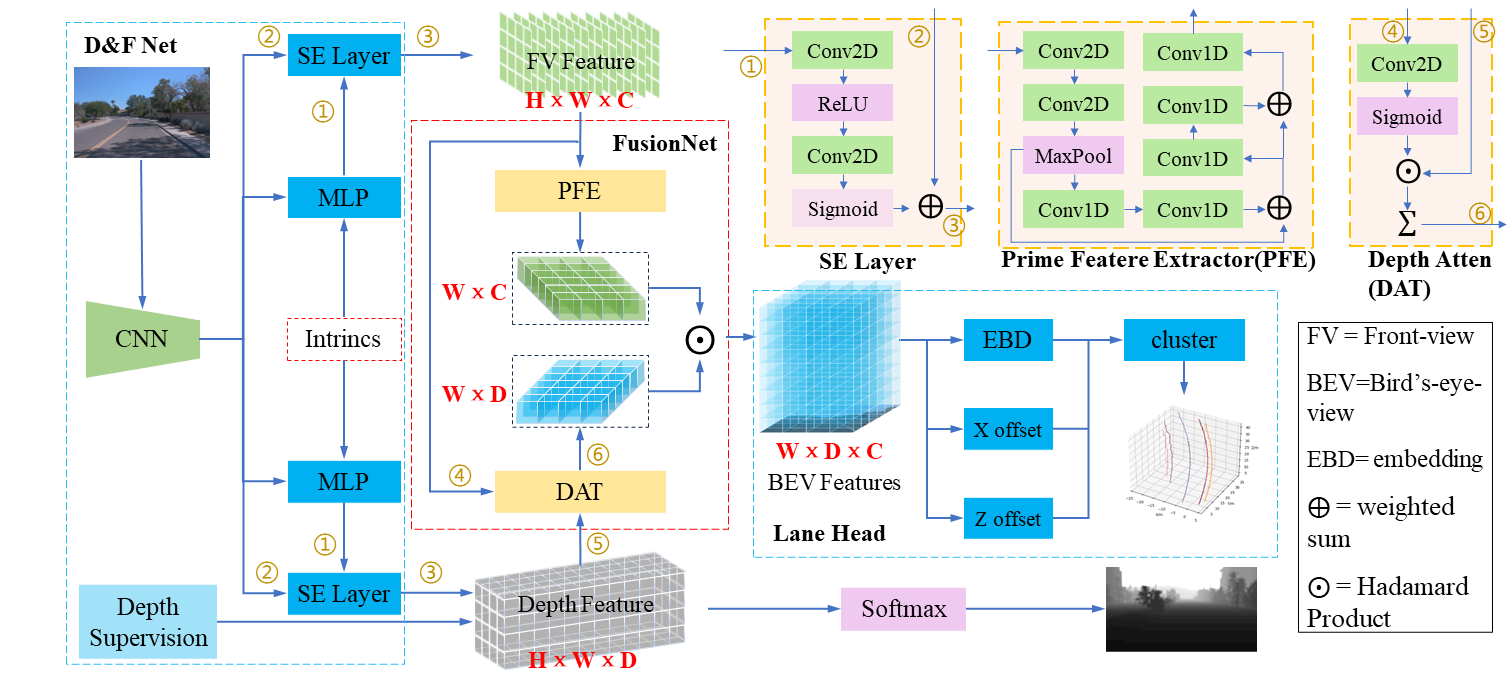}
	% 	}}
	\caption{The  overall architecture of our  proposed method DB3D-L. There are there main modue, D\&F Net, Fusion Net, and Lane Head. In  D\&F Net, SE Layer is used to realize space cross attention, in Fusion Net, PFE and DAT is used to reduce the height dimension of the two types features.}
	\label{fig:pipeline}
\end{figure*}

As a result, the precision of depth estimation is crucial  in the workflow. A small yet is designed efficient Depth Net to balance precision and efficiency, resnet is utilized as backbone, then a SE layer is intregrated to make the network has the ability of spactial attention. The purpose of the SE module is to assign distinct weights to various positions in the image within the channel domain, utilizing a weight matrix. This aims to extract more crucial feature information. As shown in Fig. \ref{fig:pipeline}, we add the camera intrincs into a multilayer perceptron (MLP) as the weight, and the outputs of Depth Net is supervised by depth truth of size $H_I \times W_I \times D$, where, at specific heights and widths, it yields $ D $ probability values that add up to 1.. Given depth truth $\boldsymbol{Dt}$, the depth loss is calculated as 

\begin{equation}
	{L_d} = \sum\limits_i^{{H_I}} {\sum\limits_j^{{W_I}} {\left\{ {\left| {{\boldsymbol{Dt}_{ij}} - {\boldsymbol{D}_{ij}}} \right| \odot {\boldsymbol{Dt}_{ij}}} \right\}} } 
\end{equation}

Inspired by muli-task learning methods \cite{dai2019hybridnet,wu2022yolop}, we shared the backbone by Depth Net and FV fratures extraction net, named D\&F Net. The FV feature and depth feature pass the common resnet and separate at the SE layer, different results are produced under different supervision.

\subsection{Fusion Net}

\begin{figure}[t]
	\centering
	\includegraphics[width=\columnwidth]{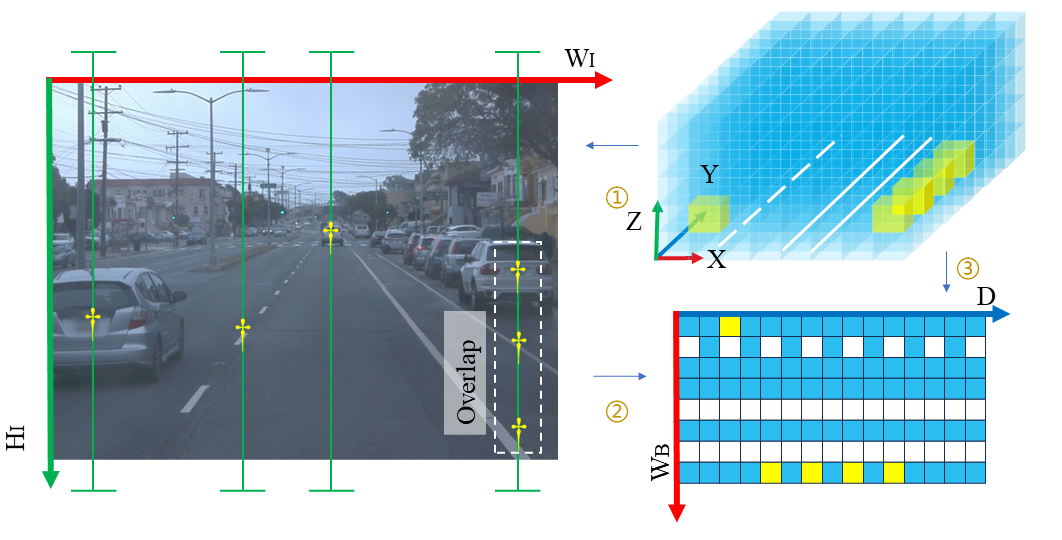}
	\caption{Image space representation, pseudo point clouds and BEV space representation. There are key elements overlap in height dimension in image space but no overlap in others. $\dagger$ represents key element. 1 represents Perspective. 2 represents IPM. 3 represents orthographic projection from top.}
	\label{fig:reduce}
\end{figure}

As shown in Fig. \ref{fig:reduce}, in image space, due to the perspective geometry, key information from farther and more outward positions may project onto the same column in the image as key information from closer positions, causing overlap in $ H $ dimension, but in the pseudo point clouds of 3D space in real sence, there are no key elements overlap, and will not overlap in BEV space. Besides, In the pseudo point cloud information recovered from the image, elevated positions in height often constitute invalid information for road scene perception, especially for traffic participant recognition and lane detection.

Recognizing the uniqueness of key information in the height channel in both 3D and BEV spaces, and aiming to avoid computations on ineffective information from elevated areas, the height channels of the two extracted features are compressed into single channels before entering the fusion module.

To reduce FV feature in size $H \times W \times C$, we adopt the Prime Feature Extraction modle (PFE) in FusionNet, PFE predicts prime feature attention for each direction (acolumn of the tensor), which consists of column-wise max-pooling, and convolutions for refinement, as shown in Fig. \ref{fig:pipeline}. By using PFE, invalid information in the height channel is filtered out and the prime FV feature in size $1 \times W \times C$ is extracted. To reduce depth feature in size $H \times W \times D$, a Depth Attention module (DAT) is designed, which learns the key information guided by the FV feature. The FV feature first goes through a convlusion layer and is activated by Sigmoid function, then it is used to do a Hadamard Product with the depth feature as a patial weight. The DAT realizes the function of spacial cross attention module, to let the depth be spatial-aware. Then we extract prime depth feature in size $D \times W \times 1$ by selecting max value for each colum in which the $ H $ dimension reduce to single.

\begin{figure}[t]
	\centering
	\includegraphics[scale=0.3]{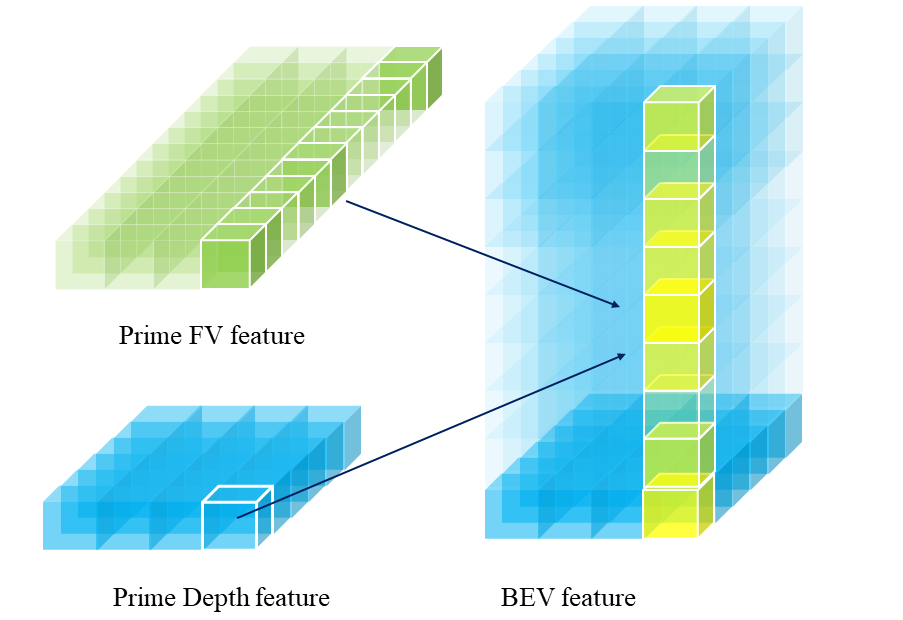}
	\caption{The process of fuse BEV frature from prime FV feature and prime depth feature.}
	\label{fig:inllustrate}
\end{figure}

The fusion module is aim to creat the BEV feature in size $D \times W \times C$ from the prime FV feature in size $1 \times W \times C$ and the prime depth in size $D \times W \times 1$.  As illustrated above, $ D $ is the probability distribution for each width. Inspired by cross attention method, for each width, the prime feature is allocated to each depth position, as a result, prime feature is distributed by the depth probability, as shown in Fig. \ref{fig:inllustrate}. With this cross-attention mechanism, features receive more emphasis in regions where the depth probability is higher. Assuming $\boldsymbol{F}$ be prime feature, $\boldsymbol{X}$ be prime depth, the BEV feature $\boldsymbol{B} \in  \mathbb{R}^{ W \times D \times C}$ is calculated as

\begin{equation}
 	\bf{B}=\bf{X}\odot\bf{F} 
\end{equation}

\noindent where $\odot$ represents Hadamard Product.

%\vspace{-1.1em}
%\begin{algorithm}[htbp]
%	\SetAlgoLined
%	\caption{Significant Overlap Labeling}
%	\KwIn{Number of NN search $k$, distance threshold $d$, ratio threshold $\tau$, point cloud $\boldsymbol{Q}$,
%	 number of divisions in the horizontal coordinate $h$,
%	 point with evaluation $\boldsymbol{p}$ in $\boldsymbol{P}$}
%	\KwOut{Return $True$ if $\boldsymbol{p}$ is significant overlapping}

%	$\boldsymbol{coor}, \boldsymbol{dist} \gets$ nearestKSearch($\boldsymbol{p}$, $\boldsymbol{Q}$, $k$) \\
%	\eIf{min$(\boldsymbol{dist})<d$}{
%		 $\boldsymbol{num} \gets$ count the number of points in $h$ regions \\
%		 \For{$i\gets1$ \KwTo $h$}{
%		 \For{$j\gets1$ \KwTo $h$}{
%		 	$ratio \gets \boldsymbol{num}\left[ i \right] /\boldsymbol{num}\left[ j \right] $\\
%		 	\If{$
%		 		(ratio<1/\tau)  \lor  (ratio>\tau) 
%		 		$}{\KwRet{False}}
%		 }}
%	 \KwRet{True}
%	}
%	{\KwRet{False}}
%	\label{alg:label}
%\end{algorithm}
%\vspace{-1.3em}

\subsection{3D Lane Detection Head}

To represent the 3D coordinates of lane lines, it is usually using discrete 3D points, rather than a set of parameters of 3D curve as the fitted curve with inevitably errors. In BEV space, the 3D coordinates can be presented as its top view projection along with the height value of each point effectively, which is a natural advantage of BEVs.

The 3D lane detection head needs to output the spatial distribution of each lane line in BEV space, along with the offset at key points. The spatial coordinate system is defined as right-front-up, as shown in \ref{fig:reduce}, and the lateral and height offsets are typically computed at a fixed longitudinal distance. The range distribution of lane lines in BEV space can be determined by a starting point plus angle offset \cite{guo2020gen}, or by using multiple BEV layers $\boldsymbol{B} \in  \mathbb{R}^{L \times W \times D \times 2}$, corresponding to a number of $ L $ lane lines. However, these approach is inefficient. Inspired by instance segmentation tasks \cite{wang2018lanenet,bolya2019yolact}, this paper divides the output BEV results into a grid of $D \times W$ and directly predicts confidence and classes for each BEV grid cell, determining whether it belongs to a particular lane line, and then predicts lateral and height offset for each cell. 
%For a cell $ b_ij $ predicted as lane line, its coordinates in 3D space are calculated as

%\begin{equation}
%	\begin{array}{c}
%		x = x_0 + i \times \Delta x  + offsx \times \Delta x, \\
%		y = j \times \Delta y,\\
%		z = offsz \times \Delta z
%	\end{array}
%	\label{eq:XP}
%\end{equation}

Therefore, the lane detection head designed in this paper needs to accomplish three tasks: a classification task to predict the category of each cell (N lane lines and background), and two regression tasks to output the X and Z offsets of the lane line points within the cell.

Finally, a fast unsupervised clustering method is used to obtain the instance-level lanes, which we refer to Post-processing algorithm in BEV-LaneDet \cite{wang2023bev}.

\subsection{Loss Design}

Fistly, we introduce a binary cross entropy loss to indicate the confidence level that each BEV grid represents a lane line point

\begin{equation}
{L_{conf}} = \sum\limits_i^{D \times W} { {{{ p}_i}\log {\hat p_i} + \left( {1 - {{ p}_i}} \right)} } \log \left( {1 - {\hat p_i}} \right) 
\end{equation}

\noindent where $\hat{p_i}$ denotes the probability of the confidence predicted by the model, and $p_i$ denotes the ground truth of confidence.
Then we classify each lane line by a instance segmentation approach \cite{wang2018lanenet}. A clustering method is used to predict the variable number of lanes. The variance of cell embeddings within the same lane and the cumulative distance from different lanes are computed. The loss is 

\begin{equation}
{L_{inst}} = {L_{{\mathop{\rm var}} }} + \frac{1}{{{L_{dist}}}}
\end{equation}

MSE loss is utilized to express the two types offsets loss. Only offsets for grid cells with a positive ground truth of confidence are calculated. For height offset, it is calucated as

\begin{equation}
% MathType!MTEF!2!1!+-
{L_{offh}} = \sum\limits_i^{D \times W} {\left| {{\hat h_i} - {{ h}_i}} \right|} {\xi _i}, \, {\xi _i} = \left\{ {\begin{array}{*{20}{c}}
		1 ,&{{ p}_i} > \sigma \\
		0 ,&else
\end{array}} \right.
\end{equation}
%$$
%L_t=\lVert \boldsymbol{t}-\boldsymbol{\hat{t}} \rVert _2, \quad    L_r=\lVert \boldsymbol{\hat{R}}^{\top}\boldsymbol{R}-\boldsymbol{I} \rVert _2, \eqno{(7)}
%$$

\noindent where $\hat h_i$ denotes height offset from prediction for each cells, and and ${h_i}$ denotes ralated offset from ground truth, $sigma$ is the threshold to determine whether a cell is a lane line.

Similarly,the x offset is calculated as

\begin{equation}
	% MathType!MTEF!2!1!+-
{L_{offx}} = \sum\limits_i^{D \times W} {\left| {\delta {\hat x_i} - \delta {{ x}_i}} \right|} {\xi _i}, \, {\xi _i} = \left\{ {\begin{array}{*{20}{c}}
		1,&{{ p}_i} > \sigma \\
		0,&else
\end{array}} \right.
\end{equation}

Note that we calculate the offset of lane line points relative to the left boundary of each BEV grid, ranging from 0 to 1, if the grid is classified as belonging to a certain lane line.

The total loss includes depth loss and 3D lane losses, which is

\begin{equation}
	% MathType!MTEF!2!1!+-
\begin{array}{c}
	L_{total} = {\lambda _d}{L_d} + {\lambda _{conf}}{L_{conf}} + {\lambda _{inst}}{L_{inst}}\\
	+ {\lambda _{offsx}}{L_{offsx}} + {\lambda _{offsz}}{L_{offsz}}
\end{array}
\end{equation}

\begin{table*}[htbp]
	%\small
	\caption{EVALUATION RESULTS ON APOLLOSIM DATASET}
	\label{table:apollo}
	\setlength{\tabcolsep}{4.6mm}
	\begin{center}
		\begin{tabular}{c|cc|cc|cccc}
			\toprule[1.5pt]
			\multirow{2}{*}{\textbf{Sence}} & \multirow{2}{*}{\textbf{Method}} & \multirow{2}{*}{\textbf{Class}} & \multirow{2}{*}{\textbf{F1}(\%)} & \multirow{2}{*}{\textbf{Acc}(\%)} 
			&\multicolumn{2}{c}{\textbf{X error }(m)}  &\multicolumn{2}{c}{\textbf{Z error }(m)} \\
			\cline{6-9}
			&{} &{} &{} &{} &near &far &near &far \\
			\midrule[1pt]
			\multirow{9}{*}{Balanced Scene} 
			&3D-LaneNet\cite{garnett20193d}	&IPM	&86.40	&89.3	&0.068	&0.477	&0.015	&0.202 \\
			&Gen-laneNet\cite{guo2020gen}	&IPM	&88.10	&90.1	&0.061	&0.496	&0.012	&0.214 \\
			&Persformer\cite{chen2022persformer}	&ViT	&92.9	&- &0.054	&0.356	&0.01	&0.234 \\
			&Anchor3D\cite{huang2023anchor3dlane}	&ViT	&95.6	&97.2	&0.052	&0.306	&0.015	&0.223 \\
			&CurveFormer\cite{bai2023curveformer}	&ViT	&95.8	&97.3	&0.078	&0.326	&0.018	&0.219 \\
			&LATR\cite{luo2023latr}	&ViT	&96.8	&97.9	&0.022	&0.253	&\textbf{0.007}	&\textbf{0.202} \\
			&BEV-LaneDet\cite{wang2023bev}	&VRM	&98.02	&97.95	&0.0263	&\textbf{0.2215}	&0.0224 	&0.2047 \\
			&DB3D-L(Ours)	&LSS	&98.30	&98.47	&\textbf{0.0197}	&0.2246	&0.0190 	&0.2244 \\
			&DB3D-L$ \dagger $(Ours)	
			&LSS	&\textbf{98.32}	&\textbf{98.52}	&0.0223	&0.2247	&0.0194	&0.2241 \\
			\midrule[1pt]
			\multirow{9}{*}{Rarely Observed} 
			&3D-LaneNet\cite{garnett20193d}	&IPM	&74.6	&72.0	&0.166	&0.855	&0.039	&\textbf{0.521}\\
			&Gen-laneNet\cite{guo2020gen}	&IPM	&78.0	&79.0	&0.139	&0.903	&0.030	&0.539\\
			&Persformer\cite{chen2022persformer}	&ViT	&87.5	&-	&0.107	&0.782	&0.024	&0.602\\
			&Anchor3D\cite{huang2023anchor3dlane}	&ViT	&94.4	&96.9	&0.094	&0.693	&0.027	&0.579\\
			&CurveFormer\cite{bai2023curveformer}	&ViT	&95.6	&97.1	&0.182	&0.737	&0.039	&0.561\\
			&LATR\cite{luo2023latr}	&ViT	&95.8	&97.3	&0.050	&\textbf{0.600}	&\textbf{0.015}	&0.532\\
			&BEV-LaneDet\cite{wang2023bev}	&VRM	&\textbf{98.57}	&\textbf{99.45}	&0.0395	&0.6028	&0.0442	&0.6089\\
			&DB3D-L(Ours)	&LSS	&97.5	&98.01	&\textbf{0.039}	&0.6113	&0.0402	&0.6190\\
			&DB3D-L$\dagger$(Ours)	&LSS	&97.63	&98.11	&0.0419	&0.6118	&0.0399	&0.6165\\
			\midrule[1pt]
			\multirow{9}{*}{Visual Variations} 
			&3D-LaneNet\cite{garnett20193d}	&IPM	&74.9	&72.5	&0.115	&0.601	&0.032	&0.230\\
			&Gen-laneNet\cite{guo2020gen}	&IPM	&85.3	&87.2	&0.074	&0.538	&0.015	&0.232\\
			&Persformer\cite{chen2022persformer}	&ViT	&89.6	&-	&0.074	&0.430	&0.015	&0.266\\
			&Anchor3D\cite{huang2023anchor3dlane}	&ViT	&93.6	&91.4	&0.068	&0.367	&0.020	&0.232\\
			&CurveFormer\cite{bai2023curveformer}	&ViT	&90.8	&93.0	&0.125	&0.410	&0.028	&0.254\\
			&LATR\cite{luo2023latr}	&ViT	&95.1	&96.6	&0.045	&\textbf{0.315}	&\textbf{0.016}	&\textbf{0.228}\\
			&BEV-LaneDet\cite{wang2023bev}	&VRM	&\textbf{96.9}	&\textbf{97.3}	&0.027	&0.032	&0.031	&0.256\\
			&DB3D-L(Ours)	&LSS	&96.52	&96.94	&\textbf{0.0263}	&0.3313	&0.0342	&0.2523\\
			&DB3D-L$\dagger$(Ours)	&LSS	&96.89	&97.02	&0.0286	&0.3552	&0.0369	&0.2452\\
			
			\bottomrule[1.5pt]
		\end{tabular}
	\end{center}
\end{table*}
%Sence				↑	X error (m)↓	Z error (m)↓

\section{EXPERIMENTS AND ANALYSIS}

\subsection{Implementation Details}

Experiments were performed on two 3D lane detection benchmarks: the Apollo simulation (ApolloSim) dataset \cite{guo2020gen} and OpenLane v1.2 \cite{chen2022persformer}. The ApolloSim dataset comprises 6000 samples from a virtual highway map, 1500 samples from an urban map, and 3000 samples from a residential area. It includes corresponding depth maps, semantic segmentation maps, and information about 3D lane lines. OpenLane comprises 160,000 images for the training set and 40,000 images for the validation set. The validation set consists of six different scenarios, including curve, intersection, night, extreme weather, merge and split, and up and down. The dataset annotates 14 lane categories, such as road edges and double yellow solid lanes.

\subsection{Results on Apollo simulation dataset}

The ApolloSim dataset comprises 10,000 monocular images captured in scenes with balanced, rarely observed, and visual variation characteristics. Every scene includes independent training and test sets. Our method is compared with state-of-the-art approaches to validate its performance using six widely accepted indicators. The evaluated indicators include F1-score, a critical measure of predictive performance, category accuracy, and errors in X and Z dimensions for near and far distances. These metrics are based on Gen-LaneNet \cite{guo2020gen}. For the methods selected for comparison, there are different categories, IPM represents methods based on inverse perspective mapping, ViT represents methods based on vision transformer, VRM represents methods based on view relation module, LSS represents methods based on 3D information recovery.

As depicted in Table \ref{table:apollo}, our method attains comparable performance across all metrics in datasets featuring diverse conditions. In balanced scenes, our proposed method outperforms all compared methods, achieving the highest F1-score and accuracy. Even in rare scenarios, our method consistently meets leading standards. Our method exhibits excellent robustness in visually dynamic environments. The X error in near distance for our method remains minimal across all conditions. $\dagger$ represents methods that prioritize F1-score.

\subsection{Results on OpenLane}

%OpenLane contains 160K training frames and 40K test frames, which is a large-scale real-world 3D lane dataset widely used in the field of autonomous driving.
The performance evaluation on the OpenLane dataset effectively reflects the real-world performance of the proposed method. We compare our method with state-of-the-art methods in F-score, X and Z dimension errors for near and far distance. As shown in Table \ref{table:openlane}, Our method continues to exhibit comparable performance on real-world datasets, particularly excelling in near-distance X error indicators.

\begin{figure*}[t]
	\centering
	% 	\framebox{\parbox{3in}{
	\includegraphics[scale=0.45]{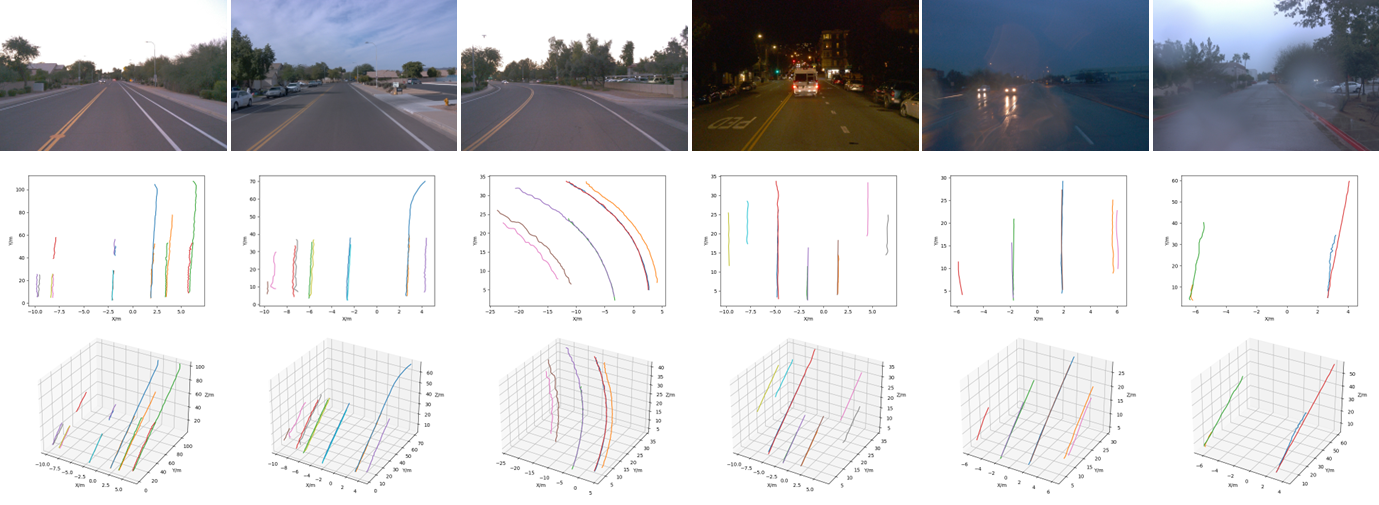}
	% 	}}
	\caption{The detection results on the OpenLane dataset, from left to right, include various scenarios such as multi-lane, occlusion, curves, night, rainy, and foggy conditions.The second row displays the detection results and ground truth in BEV space, the third row shows the detection results and ground truth in 3D space.}
	\label{fig:visualizition}
\end{figure*}

The detection results on the OpenLane dataset, depicted in Fig. \ref{fig:visualizition}, encompass diverse scenarios, including multi-lane, occlusion, curves, night, rainy, and foggy conditions. The results demonstrate that our methods successfully perform accurate lane detection in real-world scenes, exhibiting robustness in challenging lighting and weather conditions.

\subsection{Ablation studys}

The experiments in this section will be carried out on the ApolloSim, and the evaluation metrics are still based on Gen-LaneNet \cite{guo2020gen}. Our baseline uses ResNet34 \cite{he2016deep} as the backbone without depth training. We prove the effectiveness of our depth-aware methods by adding two trips: depth training, depth training with DAT. In each of these three scenarios, we train for 100 epochs separately, as shown in Table \ref{table:ablation depth}. Our based method reach 95.03\% F1-score, which is comparable in state-of-the -art methods. When depth training is used, the F1-score increases by more than two percent, the X error decreases more than 0.003m in near distance, decreases 0.02m in far distance, and the Z error  decreases 0.01m in near distance, decreases 0.04m in far distance, which proving the depth information is highly effective for accurately predicting the information of the BEV scene. When we use DAT, the optimal performance achieved, which indicates that DAT play a crucial role in the network. 

Then, An additional experiment to assess the impact of different depth training modes on performance was conducted. Three distinct deep training methods were employed . Method 1: Depth Net pre-training and then fixed backbone parameters; Method 2: Depth Net pre-training and then fixed Depth Net parameters; Method 3: Depth Net synchronous training with the model. Each method underwent training for 100 epochs

As shown in Table \ref{table:ablation depth train}. When we pretrain the depth network and fix the entire backbone, the performance of the model is considerably restricted. However, when we pretrain the Depth Net and only fix itself, the performance of the model is acceptable. It is worth noting that this method is particularly valuable in engineering applications where additional depth ground truth is lacking. Finally, when we synchronously train the depth and the entire model, the model achieves the best results.

Finally, an ablation study on FusionNet was conducted. In this experiment, a basic fusion module and the complete FusionNet were utilized independently to integrate depth information and FV features, generating BEV features, thereby validating the efficacy of our proposed FusionNet. In the basic fusion module, the height channels of depth information and FV features remained uncompressed. They were directly multiplied to produce features of size $H \times W \times D \times C$, subsequently input into a CNN to derive the final BEV features. The experimental results, presented in Table \ref{table:ablation fusion}, demonstrate a significant performance improvement when using FusionNet, coupled with a reduction in the number of parameters.

\begin{table}[t]
	\small
	\caption{Evaluation Results on OPENLANE DATASET}
	\label{table:openlane}
	\setlength{\tabcolsep}{1.8mm}
	\begin{center}
		\begin{tabular}{c|c|cccc}
		\toprule[1.5pt]
		\multirow{2}{*}{\textbf{Method}}  &\multirow{2}{*}{\textbf{F1}(\%)} 
		&\multicolumn{2}{c}{\textbf{X error }(m)}  &\multicolumn{2}{c}{\textbf{Z error }(m)} \\
		\cline{3-6}
		{} &{} &near &far &near &far \\
		\midrule[1pt]
		3D-LaneNet\cite{garnett20193d}	&44.1	&0.479	&0.572	&0.367	&0.443\\
		Gen-laneNet\cite{guo2020gen}	&32.5	&0.591	&0.684	&0.411	&0.521\\
		Persformer\cite{chen2022persformer}	&50.5	&0.485	&0.553	&0.364	&0.431\\
		Anchor3D\cite{huang2023anchor3dlane}	&53.1	&0.300	&0.311	&\textbf{0.103}	&\textbf{0.139}\\
		CurveFormer\cite{bai2023curveformer}	 &50.5	&0.452	&0.556	&0.497	&0.491\\
		BEV-LaneDet\cite{wang2023bev}	&\textbf{57.33}	&0.2623	&\textbf{0.6797}	&0.2011	&0.6229\\
		DB3D-L(Ours)	&55.24	&\textbf{0.2536}	&0.6820	&0.2017	&0.6215\\
		\bottomrule[1.5pt]
		\end{tabular}
	\end{center}
\end{table}

\begin{table}[t]
	\small
	\caption{RESULTS OF ABLATION EXPERIMENT of DepthNet}
	\label{table:ablation depth}
	\setlength{\tabcolsep}{1.8mm}
	\begin{center}
		\begin{tabular}{c|c|cccc}
			\toprule[1.5pt]
			\multirow{2}{*}{\textbf{Method}}  &\multirow{2}{*}{\textbf{F1}(\%)} 
			&\multicolumn{2}{c}{\textbf{X error }(m)}  &\multicolumn{2}{c}{\textbf{Z error }(m)} \\
			\cline{3-6}
			{} &{} &near &far &near &far \\
			\midrule[1pt]
			base(resnet34)	&95.03	&0.0307	&0.2528	&0.0497	&0.2914\\
			base+depth	&97.59	&0.0279	&0.2305	&0.03923 &0.2331\\
			base+depth+DAT	&98.15	&0.0238	&0.2111	&0.0242	&0.2101\\

			\bottomrule[1.5pt]
		\end{tabular}
	\end{center}
\end{table}

\begin{table}[t]
	\small
	\caption{RESULTS OF ABLATION EXPERIMENT OF DEPTH TRAINING}
	\label{table:ablation depth train}
	\setlength{\tabcolsep}{1.8mm}
	\begin{center}
		\begin{tabular}{c|c|cccc}
			\toprule[1.5pt]
			\multirow{2}{*}{\textbf{Method}}  &\multirow{2}{*}{\textbf{F1}(\%)} 
			&\multicolumn{2}{c}{\textbf{X error }(m)}  &\multicolumn{2}{c}{\textbf{Z error }(m)} \\
			\cline{3-6}
			{} &{} &near &far &near &far \\
			\midrule[1pt]
			Method 1	&94.92	&0.0469	&0.3113	&0.0361	&0.2383\\
			Method 2	&97.36	&0.0321	&0.2477	&0.0326	&0.2220\\
			Method 3	&98.15	&0.0238	&0.2111	&0.0242	&0.2101\\
			\bottomrule[1.5pt]
		\end{tabular}
	\end{center}
\end{table}

\begin{table}[t]
	\small
	\caption{RESULTS OF ABLATION EXPERIMENT OF Fusion Net}
	\label{table:ablation fusion}
	\setlength{\tabcolsep}{0.8mm}
	\begin{center}
		\begin{tabular}{c|c|ccccc}
			\toprule[1.5pt]
			\multirow{2}{*}{\textbf{Method}}  &\multirow{2}{*}{\textbf{F1}(\%)} 
			&\multicolumn{2}{c}{\textbf{X error }(m)}  &\multicolumn{2}{c}{\textbf{Z error }(m)} & \multirow{2}{*}{\textbf{PM}} \\
			\cline{3-6}
			{} &{} &near &far &near &far \\
			\midrule[1pt]
			\makecell{base \\ + fusion module} 	&94.92	&0.0469	&0.3113	&0.0361	&0.2383 & 684M\\
			\makecell{base \\ + FusionNet}	&98.15	&0.0238	&0.2111	&0.0242	&0.2101 & 675M\\
			\bottomrule[1.5pt]
		\end{tabular}
	\end{center}
\end{table}

%\begin{figure*}[t]
%	\centering
%	%	 	\framebox{\parbox{3in}{
		%			\includegraphics[scale=0.3]{Figure/Ablu}
		%			%			 	}}
%	\caption{The registration recall performance for different methods in ablation experiments.}
%	\label{fig_ablution}
%\end{figure*}

\addtolength{\textheight}{-0.5cm}   % This command serves to balance the column lengths

\section{CONCLUSION}

In this paper, we propose DB3D-L, a depth-aware end-to-end framework for 3D lane detection that shows comparable performance with state-of-the-arts methods. The intergrated Depth Net make it possible to do 3D recovery, a refined backbone is shared by Depth Net and feature extraction module in line of multi-task learning, PFE and DAT is designed in FusionNet to reduce the height dimension of features, and a effective fusion method is used to build BEV feature from prime FV feature and Depth frature. Extensive experiments show that DB3D-L achieves remarkable performance. We believe that our work can benefit the community and inspire further research.

\balance
\bibliography{ref}%references} %bibfile_name
\bibliographystyle{IEEEtran}

\end{document}